\documentclass[letterpaper, 10 pt, conference]{ieeeconf}

\IEEEoverridecommandlockouts 
\usepackage{amsmath}
\usepackage{bm}
\usepackage{amssymb}
\usepackage{algorithm}
\usepackage{algpseudocode}
\usepackage[sorting=none]{biblatex}
\bibliography{references}
\usepackage{comment}
\usepackage{tabularx}
\usepackage[pdftex]{graphicx}
\usepackage{subcaption}
\DeclareGraphicsExtensions{.pdf,.png,.jpeg}
\usepackage{url}
\usepackage[dvipsnames]{xcolor}

\newcommand*{\cour}{\fontfamily{pcr}\selectfont}

\algnewcommand{\parState}[1]{\State%
  \parbox[t]{\dimexpr\linewidth-\algmargin}{\strut #1\strut}}

\makeatletter
\newcommand*\titleheader[1]{\gdef\@titleheader{#1}}
\AtBeginDocument{%
  \let\st@red@title\@title
  \def\@title{%
    \bgroup\normalfont\large\centering\@titleheader\par\egroup
    \vskip1.5em\st@red@title}
}
\makeatother

\title{\LARGE \bf
Mitigating Shadows in Lidar Scan Matching using Spherical Voxels
}
\titleheader{\scriptsize This work has been submitted to the IEEE for possible publication. Copyright may be transferred without notice, after which this version may no longer be accessible.}
\author{Matthew McDermott$^{1}$ and Jason Rife$^{2}$
\thanks{$^{1}$Matthew McDermott is a student in the Mechanical Engineering Ph.D. program at Tufts University in Medford, MA. He works in the Automated Systems and Robotics Laboratory (ASAR) with Dr. Jason Rife. He received his B.S. and M.S. degrees in Mechanical Engineering at Tufts University,
        {\tt\small matthew.mcdermott@tufts.edu}}%
\thanks{$^{2}$Jason Rife is a Professor and Chair of the Department of Mechanical Engineering at Tufts University in Medford, Massachusetts. He directs the Automated Systems and Robotics Laboratory (ASAR), which applies theory and experiment to characterize integrity of autonomous vehicle systems. He received his B.S. in Mechanical and Aerospace Engineering from Cornell University and his M.S. and Ph.D. degrees in Mechanical Engineering from Stanford University.
        {\tt\small jason.rife@tufts.edu}}%
}

\begin{document}



\maketitle

\begin{abstract}
In this paper we propose an approach to mitigate shadowing errors in Lidar scan matching, by introducing a preprocessing step based on spherical gridding. Because the grid aligns with the Lidar beam, it is relatively easy to eliminate shadow edges which cause systematic errors in Lidar scan matching. As we show through simulation, our proposed algorithm provides better results than ground-plane removal, the most common existing strategy for shadow mitigation. Unlike ground plane removal, our method applies to arbitrary terrains (e.g. shadows on urban walls, shadows in hilly terrain) while retaining key Lidar points on the ground that are critical for estimating changes in height, pitch, and roll. Our preprocessing algorithm can be used with a range of scan-matching methods; however, for voxel-based scan matching methods, it provides additional benefits by reducing computation costs and more evenly distributing Lidar points among voxels.

    
\end{abstract}

\section{Introduction}

Lidar Odometry is a dead reckoning technique in which sequences of Lidar range scans are registered on top of each other to estimate vehicle movement. In contrast with sensing modalities like GNSS and INS, the accuracy of a Lidar scan match (and thus the resulting odometry estimate) is a strong function of both sensor quality and the geometric properties of the Lidar's immediate surroundings. 
In this paper we propose a new approach to mitigate shadows, which occur when regions of the surroundings are occluded (shadowed) by a closer object. The edges of the shadowed regions move when the Lidar moves. Moving shadows violate the assumption that the surroundings are static and, thereby, cause systematic biases in the scan-matching process. 

Broadly speaking, existing scan-matching methods fall into one of three categories:  end-to-end machine-learning (ML) methods that directly relate two point clouds, feature-based methods that classify groups of points as geometric structures recognizable across point clouds, and voxel-based methods that cluster points via a grid to enable alignment between clouds.  In concept, our shadow removal method can be applied as a pre-processing step for any of these methodologies.  As such, it is instructive to provide a brief overview of each category.

End-to-end ML methods estimate transformations directly from a pair of raw point clouds, either through direct estimation \parencite{LO-Net} or by projecting the 3D point cloud to a 2D \textit{spin image} and  extracting contextual information from the scene \parencite{vertexnet}. ML methods are advantageous in that they have the potential to achieve high accuracy because they can capture subtle effects in real-world data not easily captured by simple geometric models. However, ML-based methods are potentially limited because of their unpredictable patterns of error \parencite{DNNunsafe} and because of persistent challenges in registering point clouds that are only partially overlapping \parencite{DNNsurvey}. Applying shadow-removal as a preprocessing step may help reduce the complexity of the ML-based model required to compute the transformation between two scans.

Feature-based scan-matching methods identify structures in the point cloud and match those structures between scans. The simplest such approach, dubbed \textit{Iterative Closest Point} (ICP), simply matches each individual point between two scans \parencite{Besl}. Early ICP algorithms are quite computationally intensive because of the large number of possible correspondences, so ICP variants have been developed that accelerate processing by limiting correspondence to a local neighborhood \parencite{octreeICP}, by down-sampling the number of points used in each scan  \parencite{ProbabilisticICP,ML_PC_simplification}, or by improving initialization  \parencite{RGBDICP,Kassas}. Other feature-based methods identify and correspond edges \parencite{LOAM,lego}, planes \parencite{GICP}, or regions of similar Lidar reflectivity \parencite{iloam} between scans. Mixed methods also exist that extract then match features using an ML-based approach, such as a deep neural-network \parencite{DMLO}. Shadow removal has the potential to aid feature-based methods by ensuring that shadow edges are not labeled as valid features.

Voxel-based methods subdivide the scene into a grid of adjacent volume elements (aka, \textit{voxels}). The distribution of points within each voxel is characterized, and scans are aligned to maximize the similarity of the distributions within each voxel. The best known voxel-based method is the \textit{Normal Distributions Transform} (NDT), which describes the distribution of points in each voxel using a Gaussian density function \parencite{Biber}. Unlike feature-based methods, NDT does not ascribe any specific shape to characterize surfaces, a property which makes it more robust to processing scenes with arbitrary terrain \parencite{NDTvsICP}. NDT is also known for being computationally efficient \parencite{NDTvsICP}. Another recently introduced voxel-based approach is the \textit{Iterative Closest Ellipsoidal Transformation} (ICET), a method notable for analytically predicting solution accuracy as a function of scene geometry \parencite{ICET,3DICET}. ICET also includes a capability to recognize and exclude point groupings compromised by  \textit{Aperture Ambiguity} \parencite{shimojo1989occlusion}. Shadow removal can benefit  voxel-based scan matching by  mitigating systematic errors that corrupt voxel point distributions. As we will show, shadow mitigation can also reformulate the voxel grid for more efficient and accurate computations. 

The key contributions of this paper are to introduce a novel shadow-removal method, which acts as a preprocessing step before lidar scan-matching, and to characterize the benefits of the shadow-removal method via simulation. In our performance evaluation, we focus on voxel-based scan matching, and in particular on ICET. Because ICET predicts the error covariance of the output states, it is possible to demonstrate clearly that shadow mitigation is effective if simulations show the predicted accuracy matches the true accuracy after shadow mitigation is applied (but not before). In the following sections, we will explain the shadowing error mechanism, detail our shadow-removal approach, and evaluate its performance using high-fidelity simulations.


\section{Range Shadowing Effect}
\begin{figure}[t]
\centering
\includegraphics[width=3.3in]{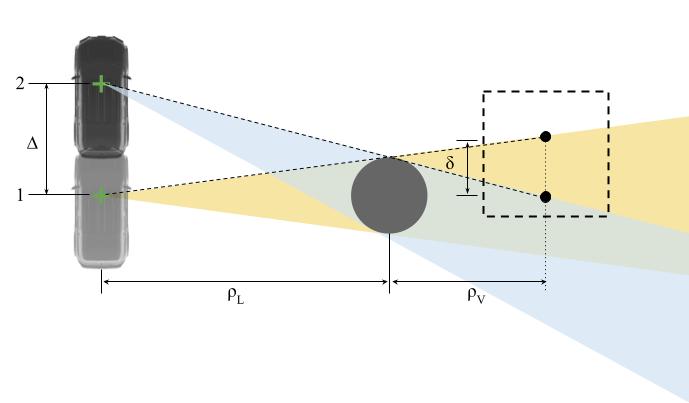} 
\caption{Schematic of shadows created by a moving Lidar (Vehicle graphic credit: ShapeNet \parencite{shapenet})}
\label{fig:shadow}
\end{figure}

To help explain the impact of range shadowing on voxel-based scan matching, this section provides an illustrative geometric model. Consider the scenario shown in Fig. \ref{fig:shadow}. The Lidar moves with a vehicle through a distance $\Delta$ from one position to another (from the cross labeled 1 to the cross labeled 2). The Lidar beam intersects a column (shown as a circular cross-section in the figure) at a distance $\rho_L$ from the Lidar. The column casts a shadow, which extends through many voxels including a voxel of interest (dashed rectangle). The voxel is a distance of approximately $\rho_V$ from the column. 

As the Lidar moves, the edge of the shadow shifts by a distance $\delta$ (which we measure parallel to $\Delta$). The Lidar beam and shadow edge are tangent to the column at a point that moves little in response to Lidar movement, assuming the radius of the column is small compared to $\rho_L$. For this reasonable case, we can use similar triangles to give:

\begin{equation}\label{eq:similar}
    \delta=\frac{\rho_v}{\rho_l}\Delta
\end{equation}

As the Lidar moves (toward the top of the page), the edge of the shadow within the voxel moves in the opposite direction, toward the bottom of the figure.  This decreases the size of the shadowed region and increases the width of the point cloud in the voxel. Algorithms like NDT and ICET track the mean location of points within the voxel, so as the shadow recedes, the mean shifts down by $|\Delta\bar{x}|$. Using a one-dimensional, uniform-density approximation for the first moment, the change in the mean is:

\begin{equation}\label{eq:com1}
    |\Delta\bar{x}| = \frac{1}{2}\delta
\end{equation}


\noindent 
Combining (\ref{eq:similar}) and (\ref{eq:com1}) gives:


\begin{equation}\label{eq:com2}
    |\Delta\bar{x}| = \frac{\rho_v}{\rho_l}\frac{\Delta}{2}
\end{equation}

\noindent According to this equation, the point cloud has an apparent shift $|\Delta\bar{x}|$ equal to half $\Delta$, scaled by a distance ratio. The distance-ratio is $\rho_v/\rho_l\in[0,\infty)$, which is zero just behind the column and increasingly large farther behind it. 
%
(We assume the Lidar motion is small enough that the shadow edge always crosses through the voxel.) 

From our analysis of Fig. \ref{fig:shadow}, we conclude importantly that the shadow (i) impacts many voxels in which it (ii) amplifies apparent vehicle motion. The amplification of vehicle motion occurs because, from the point of view of the moving Lidar, the column appears to shift downward (on the page) while the shadow appears to shift even farther downward. Since the shadow edge cuts through many voxels, a systematic bias is introduced into the perceived motion inferred from all of those voxels. This systematic bias affects all voxels along the edge of the shadow, and according to (\ref{eq:com2}), the  bias increases moving progressively farther away from the column.

In the literature, we observe two mechanisms that have been proposed for mitigating the effects of shadowing. The first approach is to remove all voxels on or near the ground plane \parencite{lego,squeezeseg}. This approach is a widely-used heuristic, but it comes with a cost. Removing the ground plane signficantly reduces the ability of Lidar to estimate pitch, roll, and vertical translation. Moreover, the heuristic assumes that shadows lie predominantly on a flat plane (e.g. the road), an assumption that may not be valid in some urban areas (when Lidar shadows are cast on walls) or in offroad terrain.

The second approach looks for jumps in a 1D ranging signal  \parencite{Huber}. These jumps correspond to adjacent lidar measurements that reflect from different surfaces (e.g. the column or the ground behind it) and aid in detecting the edge of a shadow. Because these methods only apply to 1D line scans, our new approach, introduced in the following section, adapts the jump-detection concept to apply to 2D  scans (e.g., sampled over both azimuth and elevation).

\section{Voxel-based Jump Detection}
In this section, we apply a spherical-gridding approach in a novel way, to mitigate Lidar shadows. Our approach operates before each scan match, excluding problematic points from the primary scan and the secondary scan, which is transformed to align with the primary.
\subsection{Primary Scan}
It is straightforward to convert a points from a Cartesian vector $\mathbf{q}$ to a spherical form with radius $r$, azimuth $\alpha$, and elevation $\beta$. They are related by:

\begin{equation}
    \label{eq:coordTrans}
    \mathbf{q} = 
    \left[\begin{array}{cc}
         r \cos(\alpha)\cos(\beta)\\
         r \sin(\alpha)\cos(\beta)\\
         r \sin(\beta)\\
    \end{array}\right]
\end{equation}

\begin{figure}[b]
\centering
\includegraphics[width=2in]{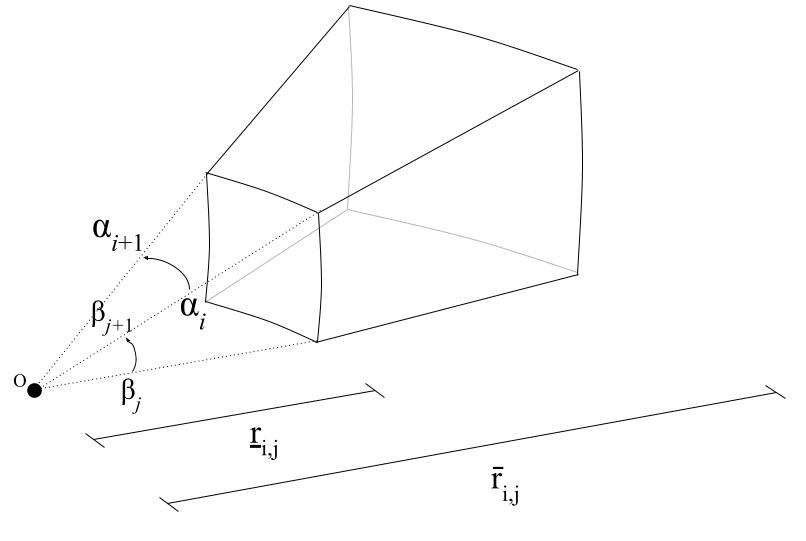}
\caption{Voxel boundaries}
\label{fig:voxel}
\end{figure}

\noindent If the Lidar unit is modeled as a point, the spherical description is useful because each beam emanates radially outward along a particular azimuth $\alpha$ and elevation $\beta$. In a spherical grid, Lidar beams do not cross grid boundaries, so it is easy to recognize the nearest radial object and exclude any objects behind it, which are likely to be shadowed. By extension, it is not necessary to fully populate the spherical grid with voxels in the radial direction, since we need only consider the voxel(s) associated with the nearest object. In fact, it is convenient to define only one radial voxel for each azimuth and elevation direction, where the radial limits are adapted to fit the nearest object. This concept for a voxel with adaptive radial boundaries is illustrated in Fig. \ref{fig:voxel}; an algorithm for obtaining the adaptive boundaries is described below and summarized as Algorithm \ref{tab:algo}.

In constructing our voxel grid, we start with a set of wedge-shaped voxels (no radial limits) indexed by  lower azimuth and elevation limits $\alpha_i$ and $\beta_j$, respectively.  If the Lidar points in scan $K$ are indexed $k \in K$, then each point is characterized by the coordinates ${}^{(k)}r, {}^{(k)}\alpha$, and  ${}^{(k)}\beta$. For voxel $(i,j)$, define the set of radial coordinates to be $V_{i,j}$:

\begin{equation}
    \label{eq:Vij}
    \begin{gathered}
        V_{i,j}=\{{}^{(k)}r\} \quad \forall \, k \quad \text{  s.t.} \\
        {}^{(k)}\alpha \in [\alpha_i,\alpha_{i+1}) \\
        {}^{(k)}\beta \in  [\beta_j,\beta_{j+1})
    \end{gathered}
\end{equation}

\begin{algorithm} [b]
    \begin{algorithmic}[1]
    \caption{Adaptive Radial Boundaries}
    \label{tab:algo}

    \State Initialize point cloud \textbf{K}
    \State Transform all points $k$ to spherical coordinates with (\ref{eq:coordTrans})
    \For{each wedge-shaped voxel $(i,j)$}
        \State Extract radii to obtain $V_{i,j}$ with (\ref{eq:Vij})
        \State Sort $V_{i,j}$ (ascending) and assign indices $l \in [0,L]$ 
        \State Set default index of inner bound to $l_{min} = 0$
        \State Set default index of outer bound to $l_{max} = L$
        
        \For{each ordered point $l$ (after point 0)}
            \State compute difference $e_l = r_{i,j}(l)-r_{i,j}(l-1)$
            \If{jump detected ($e_l > T$)}
                \If{sufficient points ($l-l_{min} > N)$}
                    \State reset outer bound to $l_{max} = l-1$
                \Else
                    \State reset inner bound to $l_{min} = l$
                \EndIf
            \EndIf
        \EndFor
        \State Set $\underbar{r}_{i,j} = r_{i,j}(l_{min})$ and $\bar{r}_{i,j} = r_{i,j}(l_{max})$ 
        \State Exclude all points with $l<l_{min}$ or $l>l_{max}$
\EndFor
\end{algorithmic}
\end{algorithm}

Our approach to segmenting the nearest object is to add an outer radial limit $\bar{r}_{i,j}$ if a radial jump between scan points is detected. This generalizes the 1D jump-detection methods of \parencite{Huber} to a spherical grid. An inner radial bound $\underbar{r}_{i,j}$ is also introduced, to screen out low-density stray points that may appear near the Lidar unit. Note that the algorithm depends on two parameters, the jump-distance threshold $T$ and the minimum number of points $N$ that define a valid object.

\subsection{Secondary Scan} 
Scan matching methods compute a rigid transformation that aligns (or \textit{registers}) a secondary scan to the primary. In general, the shadow-mitigation process of Algorithm \ref{tab:algo} is intended as a preprocessing step for both scans. In voxel-based methods, however, the benefits of shadow-matching can be attained by applying the algorithm only to the primary scan. Since a common voxel grid is used for both scans, since points for either scan are only analyzed if they fall into a voxels, and since the voxels exclude shadowed regions of the scene, then the shadow-mitigation benefits established for the first scan transfer to the second through their shared voxelization. This simplification gives a slight efficiency benefit with no apparent impact on accuracy, so we adopt it for our simulations described below.


\subsection{Iterative Scan Matching}
Once shadow mitigation has been applied, the rigid transformation relating the scans can be estimated using existing scan matching techniques. For compatibility with existing algorithms, we  frame the rigid transformation in Cartesian coordinates, in terms of a rotation matrix $\mathbf{R}$ and a Cartesian translation through the scalars $\{x, y, z\}$. For each point $k$ in the secondary scan, the rotation and translation can be applied to map the second-scan location ${}^{(k)}\textbf{p}$ to its equivalent location ${}^{(k)}\textbf{q}$ in primary-scan coordinates:

\begin{equation}\label{eq:twoDtransform}
    ^{(k)}\textbf{q} = \textbf{R}
    {}^{(k)}\textbf{p} - 
    \begin{bmatrix}
        x &
        y &
        z 
    \end{bmatrix}^T
\end{equation}

For voxel-based scan-matching methods like NDT and ICET, Algorithm \ref{tab:algo} both eliminates shadowed regions of the scene and defines a useful voxel grid. Any points from the primary or secondary scan whose locations ${}^{(k)}\textbf{q}$ fall outside the defined voxels are excluded from analysis. Point distributions are then compared for each voxel that contains a sufficient number of points from both scans, in order to compute the rotation matrix \textbf{R} and the translation $\{x,y,z\}$.

\begin{figure*}[h]
\centering
\begin{subfigure}[b]{0.34\textwidth} 
    \includegraphics[width=\textwidth]{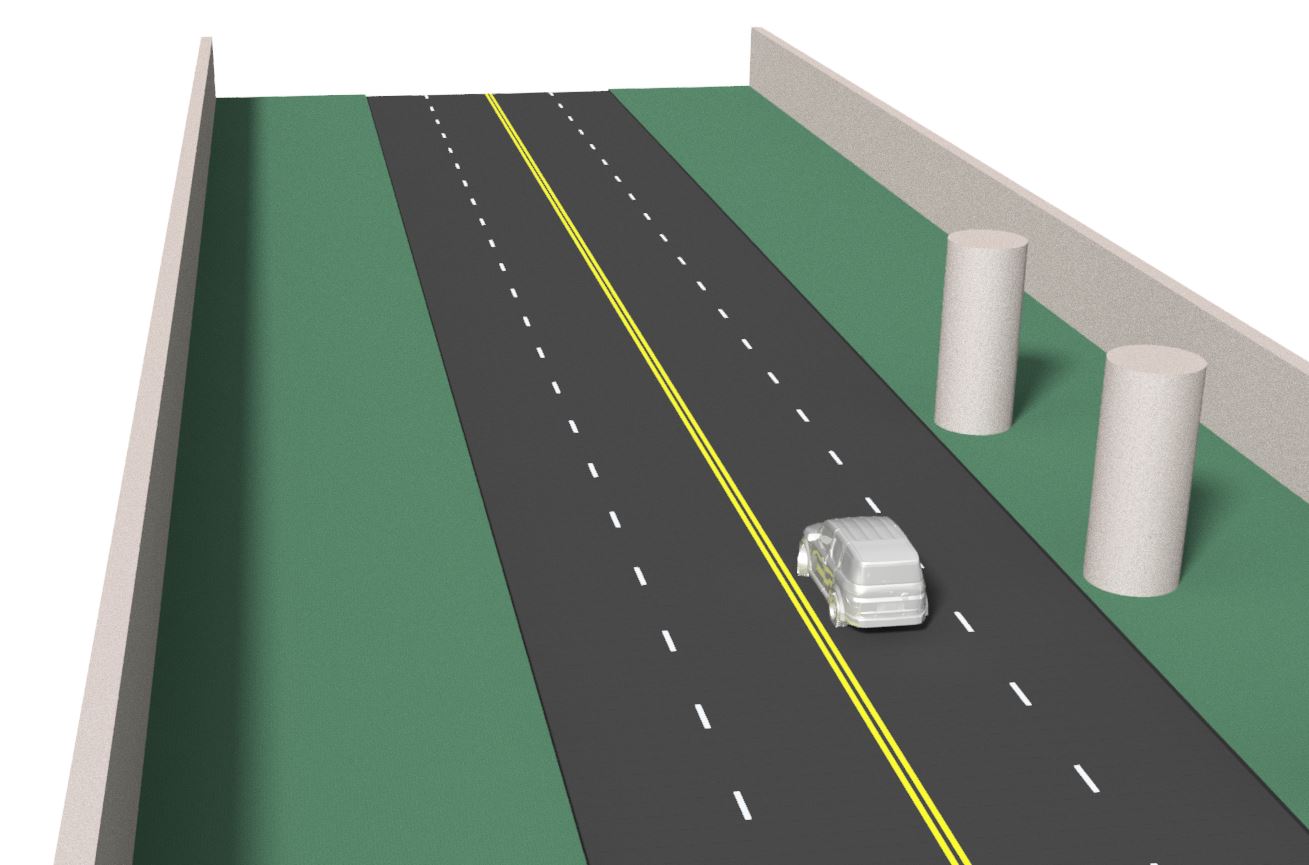} 
    \caption{Rendered scene}
    \label{fig:scene1}
\end{subfigure}
\hskip 5pt
\begin{subfigure}[b]{0.29\textwidth}
    \includegraphics[width=\textwidth]{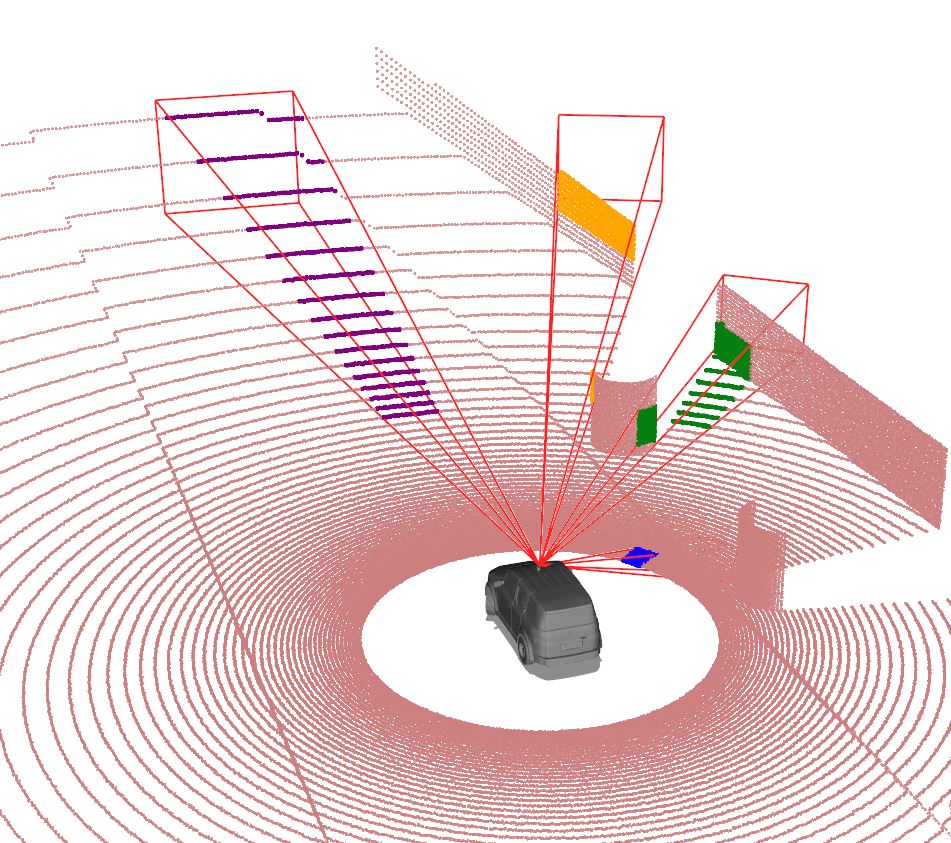}
    \caption{Wedge-shaped bins}
    \label{fig:radial_bin}
\end{subfigure}
\hskip 5pt
\begin{subfigure}[b]{0.29\textwidth}
    \includegraphics[width=\textwidth]{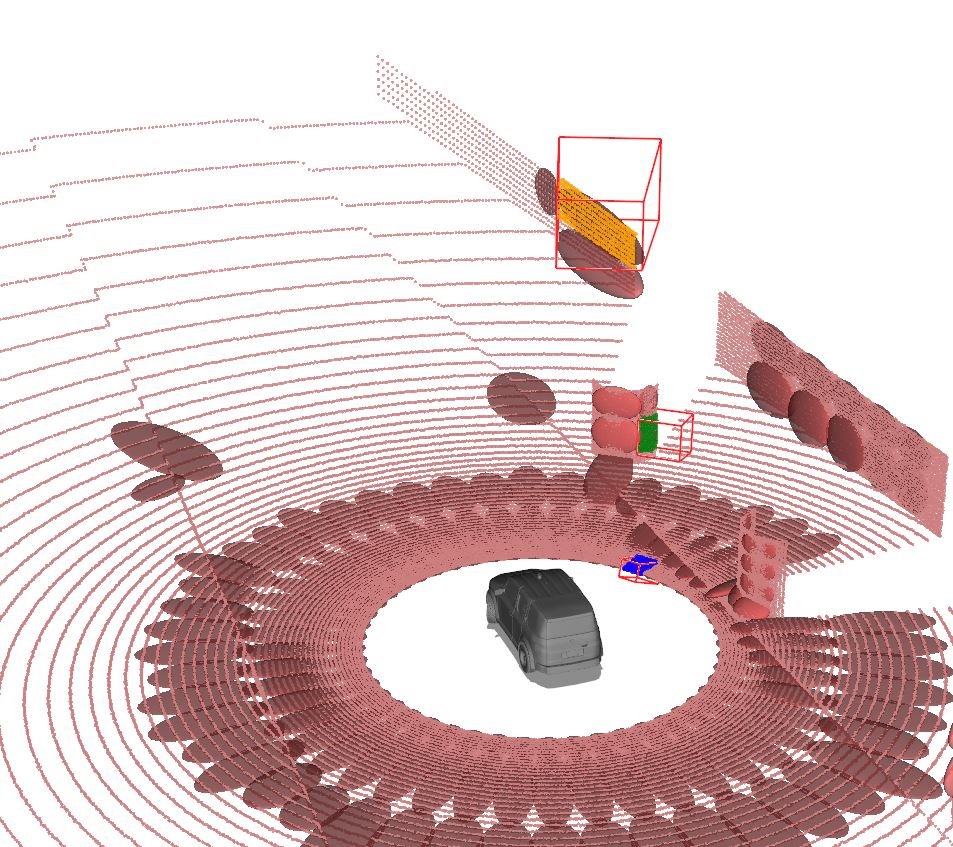} 
    \caption{Truncated voxels}
    \label{fig:full_cells}
\end{subfigure}
\vskip 10pt
\begin{subfigure}[H]{0.8\textwidth}
    \includegraphics[width=\textwidth]{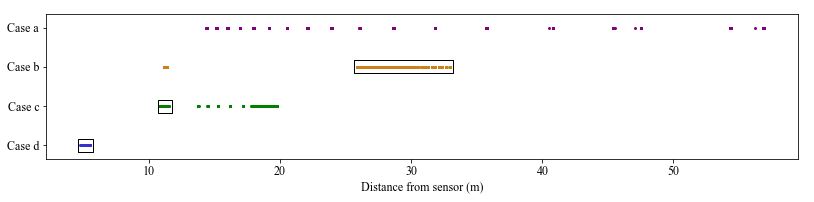}
    \caption{Radial distribution of points wedges in Fig. \ref{fig:radial_bin}}
    \label{fig:radial_dist}
\end{subfigure}
\caption{Simulated four-lane highway}
\label{fig:combinedScene1}
\end{figure*}

\section{Illustrative Example}
In order to better understand the proposed shadow-mitigation algorithm, it is helpful to explore an example in detail. Consider the simulated scene shown in Fig.~\ref{fig:combinedScene1}, where a vehicle using Lidar for navigation (the \textit{ego-vehicle}) passes a series of round columns that occlude portions of the ground plane and wall. The figure shows four views of the same scene. A visual rendering of the vehicle and scene are shown in Fig.~\ref{fig:scene1}. Simulated Lidar scan points observed as the vehicle passes the last two columns are shown in  Fig.~\ref{fig:radial_bin}; the same image also shows four wedge-shaped bins (in selected directions). The voxels generated by Algorithm~\ref{tab:algo} are shown in Fig.~\ref{fig:full_cells}; the same image also shows the covariance ellipsoids describing the distribution of points in those voxels. In moving from Fig.~\ref{fig:radial_bin} to Fig.~\ref{fig:full_cells}, it is clear that Algorithm~\ref{tab:algo} prunes voxels in some directions. In order to understand the pruning process (and the process of determining the inner and outer radial bounds for each voxel), Fig.~\ref{fig:radial_dist} visualizes the point distributions in the four wedges identified in Fig.~\ref{fig:radial_bin}.

The four wedges were selected to highlight different aspects of how Algorithm~\ref{tab:algo} functions. The wedges are labeled \textit{case a} through \textit{case d}. In \textit{case a}, marked with purple, the points lie on parallel scan lines crossing the ground plane. Because they are separated by equal intervals in elevation angle, scan lines are separated by an increasing distance along the ground moving toward the horizon.  The scan lines are never close enough to accumulate $N$ points without a radial distance jump larger than the threshold $T$. As a consequence, all of the points in the wedge are excluded, so no associated voxel appears in Fig.~\ref{fig:full_cells}. In \textit{case b}, marked with yellow, a very small cluster of points occurs at a short distance (about 12 m) and a much larger cluster of points begins at a farther distance (about 25 m). The nearer points are the edge of a column; the farther points are returns from an oblique section of the wall. There are too few points in the near cluster (less than $N$), so the inner and outer radial bounds are determined based on the farther cluster, with bounds indicated by the black box shown for \textit{case b} in Fig.~\ref{fig:radial_bin}. The same yellow-coded voxel appears in 3D in Fig.~\ref{fig:full_cells}. By contrast, \textit{case c}, marked with green, captures the other edge of the same column, but with more than $N$ points in the nearer cluster. As a consequence, the radial bounds of the voxel bracket the column and exclude the wall behind, as shown by the black box for \textit{case c} in Fig.~\ref{fig:radial_bin} and by the green-coded voxel in Fig.~\ref{fig:full_cells}. Finally, \textit{case d}, marked by blue, captures a section of the ground plane near to the ego vehicle. All of the blue points in the wedge shown in Fig.~\ref{fig:radial_bin} and in Fig.~\ref{fig:radial_bin} are close together, and so no points are excluded from the voxel.  The inner radial bound $\underbar{r}$ and outer bound $\bar{r}$ capture all points in the wedge, as shown by the black box for \textit{case d} in Fig.~\ref{fig:radial_bin}.

In this illustration, and in our subsequent performance evaluations, the bins are uniformly $7.2^{\circ}$ wide in azimuth and elevation,
the distance threshold is $T = 0.2$ m, and the minimum point count is $N = 50$. The four cases described above illustrate the tradeoffs in setting these parameters. The angular scan density is higher in the azimuth direction than the elevation direction, so setting the threshold $T$ is non-trivial. The selected threshold (0.2 m) is low enough to clearly distinguish most occluded objects (e.g. the wall and the column behind) and high enough to cluster points on most oblique surfaces (e.g. the ground plane near the ego vehicle and the vertical wall at the edge of the road); however, the threshold is not high enough to link points on a highly oblique and distant horizontal surface (e.g. the ground plane far from the ego vehicle). Increasing $T$ would allow us to capture more of the ground plane, but at the expense of losing the ability to exclude some shadows. 

Note the minimum point count $N$ was intentionally set to be higher than the number of points spanning a wedge along a single line scan (at fixed elevation for a rotating Lidar).  This means that a voxel is defined for a surface only if least two scan lines spaced closer than $T$ cross the wedge. If $N$ were lower, then a single scan line would be selected to define the voxel in \textit{case a}, which would result in malformed distribution ellipsoid (as compared to those shown in Fig.~\ref{fig:full_cells}, where the distribution of points in ground-plane voxels is represented by a well-defined flat, or \textit{oblate}, ellipsoid). At the same time, $N$ cannot be arbitrarily high, as the edge of a nearby object might be excluded, thus failing to remove a shadow.  As is, the nearby column edge in \textit{case b} is not recognized as an occluding surface, even though the column casts a shadow on the wall behind.  Arguably, \textit{case b} represents a rare mis-classification. In practice, we have observed our value for $N$ to be suitable, such that surfaces with fewer than $N$ points did not introduce a strong enough shadow to observably bias performance, as evidenced by the results of the next section.

An additional important consideration is that it is helpful to pad the radial limits slightly, to increase the size of each voxel. The reason is that surfaces seen in the secondary image may not align exactly with those in the primary image. For example, marginally farther points on a column or nearer points on the back wall might become visible in the secondary scan. To account for this we modify Algorithm~\ref{tab:algo} slightly to pad the cell volume. The padding reduces the inner radial limit and increases the outer radial limit by as much as 0.5 m (or half the distance to the next excluded point, whichever is less).

From a computational point of view, our gridding process decomposes the scene into a two-dimensional grid (with one voxel in each azimuth and elevation direction); working with this grid is much more efficient than working with a conventional three-dimensional Cartesian grid. This computational benefit can speed up processing or allow for a higher grid resolution for the same total processing time. Another advantage is that the number of points in each wedge-shaped bin (before shadow mitigation) is equal by design. This avoids a well-known limitation of Cartesian grids, where point density is much higher in near bins than far bins. These benefits of spherical gridding also apply as compared to cylindrical coordinates, as used in \parencite{RadarNDT}. Cylindrical coordinates, like Cartesian coordinates, do not align with Lidar beams, making shadow mitigation difficult. Also, using cylindrical coordinates creates a computationally less efficient three-dimensional grid, where point densities are nonuniform across voxels.



\section{Performance Evaluation}

\begin{figure}[b]
\centering
\includegraphics[width=0.25in]{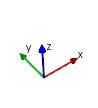}
\includegraphics[width=3.3in]{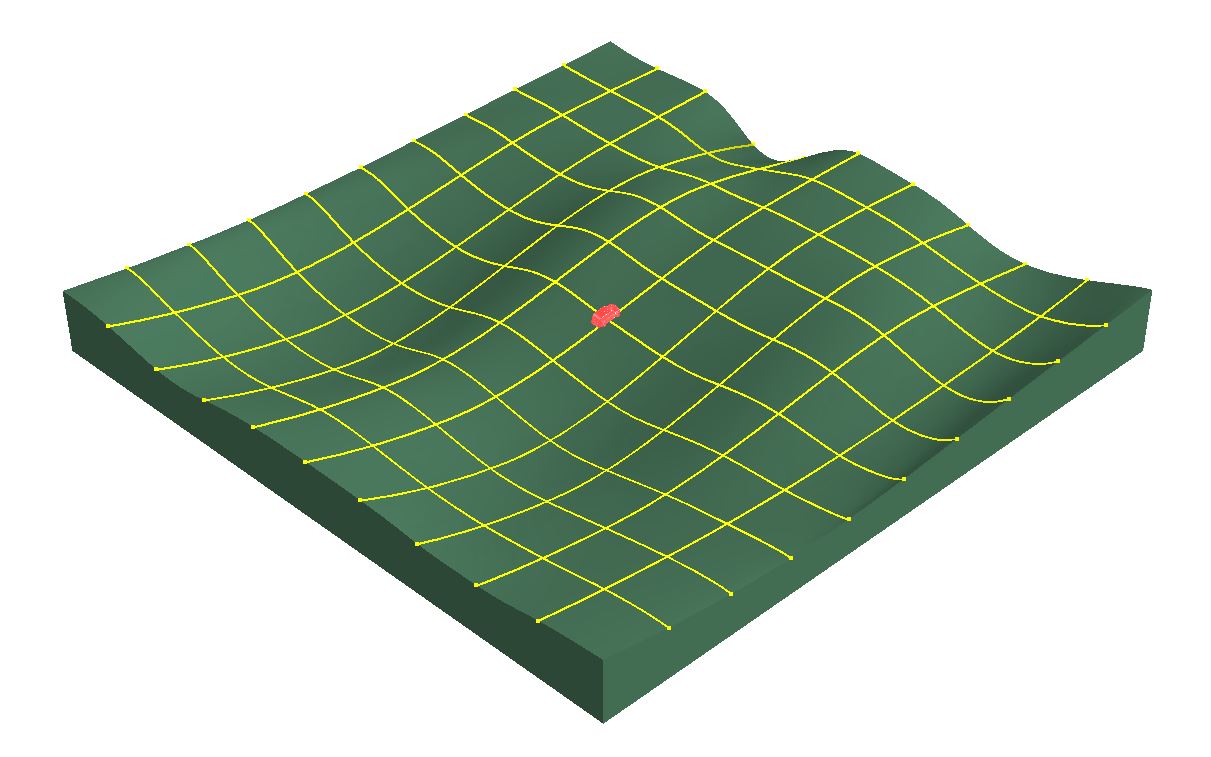}
\caption{Hilly off-road terrain (100x100m)}
\label{fig:scene2}
\end{figure}

In this section we evaluate the degree to which our shadow-mitigation methods enhance algorithm performance. To this end, we performed computations using three algorithm variations: conventional ICET with a Cartesian grid, conventional ICET with shadow mitigation via ground-plane removal, and ICET with shadow mitigation via Algorithm~\ref{tab:algo}. We compared performance using two simulated scenes. One scene was the roadway scene from Fig.~\ref{fig:combinedScene1}, featuring sound-absorption walls beside the road and a series of 10 cylindrical pillars, two of which are shown in Fig.~\ref{fig:scene1}. The second scene was an off-road terrain, shown in Fig. \ref{fig:scene2}. For each scene, multiple Monte-Carlo trials were computed with randomized Lidar errors. Since true motion is known in the simulated environment, scan-matching estimates were compared to the truth to compute an error, which was characterized statistically over the set of Monte-Carlo trials. In all, our analysis considered 120 trials on each terrain (3 samples for scan pairs in each of 40 locations along the roadway, and 6 samples for scan pairs in each of 20 locations on the off-road terrain).

Scenes were created as solid bodies in AutoDesk~Inventor and, with {\cour extendedObjectMesh} in Matlab, converted to surface triangulations. Surfaces were sampled in Matlab using {\cour monostaticLidarSensor}  to form simulated Lidar point clouds. In an attempt to maintain consistency with other benchmarks, we calibrated our virtual Lidar sensor to match the specifications of the \textit{Velodyne HDL-64E} used in the KITTI dataset \parencite{Geiger2013IJRR}. For the roadway scene, the Lidar translates 20 m along the center of its lane at a constant velocity of $5 \frac{m}{s}$. For the off-road scene, the Lidar translates at a constant forward velocity of  $5 \frac{m}{s}$ and and rotates at $30 \frac{deg}{s}$ about the vertical axis, moving for 2.1 seconds and producing 20 pairs of frames sampled at 10Hz. Scan lines were generated assuming the Lidar is on a gimball (such that it does not tilt relative to the gravity vector). In the roadway scene, the columns along the roadway cause shadowing on the ground plane and vertical walls behind; in the off-road scene, the natural contours of the hills create terrain shadows, which are notoriously difficult to mitigate \parencite{terrain_distortion}.

For performance-evaluation purposes, our shadow-matching algorithm was integrated with the ICET scan matching algorithm. ICET, as described in \parencite{3DICET}, predicts the scan error from first principles, in the absence of systematic biases. When shadow mitigation is successful, the probabilistic accuracy prediction of ICET should match the statistical accuracy obtained from the Monte Carlo simulation. By comparison, we expect the two accuracy metrics to diverge if shadow matching is not successful. For ICET analyses on Cartesian grids, cubic voxels were used, with 3m edge length (a dimension tuned for favorable performance, so as not to artificially penalize the Cartesian analysis). For the roadway scene, the voxel grid was aligned with the vertical wall passing through the middle of associated voxels (noting that Cartesian methods perform better in this configuration than when walls clip through voxel corners).

\section{Results} 

After running Monte Carlo simulations, motion was estimated with scan matching. Errors were computed in each of six directions, including the translational directions $\{x,y,z\}$ and the rotational angles $\{\phi,\theta,\psi\}$, or roll, pitch and yaw. In each case the mean of the errors was computed and found to be negligibly small. Standard deviations for each error are tabulated below for the roadway scene, in Tab.~\ref{tab:t-1} for the Cartesian grid analysis, in Tab.~\ref{tab:t-2} for the Cartesian grid analysis with ground-plane removal, and in Tab.~\ref{tab:t-3} for our proposed shadow-removal approach using spherical coordinates. Standard deviations are also tabulated for the offroad terrain scene, in Tab.~\ref{tab:t-4} for the Cartesian grid analysis and in Tab.~\ref{tab:t-5} for our proposed shadow-removal approach. Ground-plane removal is ill-defined for the offroad scene, so results are not tabulated for that case.

In nearly all trials, the ICET algorithm converged. The exceptions occurred when the Cartesian grid was used for the roadway scene (with or without ground-plane removal). Convergence failures occurred either when the algorithm diverged entirely or when it failed to escape the local minima of the initialization. These situations were easily detectable, and they were omitted from our analysis. By comparison, computations on the spherical coordinate system generated by our new algorithm successfully converged in all trials on both scenes. 

For the roadway scene, the effect of shadowing is most pronounced in the along-track (or $x$ direction). Regardless of shadowing, translational errors are larger in the $x$-direction than in other directions (by a factor of about 5 times), because there is a relative dearth of features to mark progress along the road. For the Cartesian grids (with and without ground-plane removal), shadowing increases the actual $x$-direction errors substantially above the prediction (actual $\sigma$ more than twice predicted value). By comparison, our new shadow-mitigation algorithm achieves an actual error that is very close to the predicted error in the $x$-direction (actual $\sigma$ within 2\% of the predicted value). Shadow mitigation with  

\begin{table}[H]
    \caption{Roadway - Cartesian grid (with ground plane) }
    \setlength{\tabcolsep}{0.3\tabcolsep}
    \begin{tabularx}{0.48\textwidth} { 
      | >{\centering\arraybackslash}X 
      | >{\centering\arraybackslash}X 
      | >{\centering\arraybackslash}X
      | >{\centering\arraybackslash}X
      | >{\centering\arraybackslash}X
      | >{\centering\arraybackslash}X
      | >{\centering\arraybackslash}X
      | >{\centering\arraybackslash}X | }
     \hline
      & std error $x$ (cm) & std error $y$ (cm) & std error $z$ (cm) & std error $\phi$ (deg) & std error $\theta$ (deg) & std error $\psi$ (deg) \\
     \hline
    Actual* & 0.978 & 0.0226 &  0.0197 & 0.00259 & 0.000897 & 0.00106 \\
    \hline
    Predicted & 0.467 & 0.0220 & 0.0175 & 0.00210 & 0.00125 & 0.00102\\
    \hline
    \end{tabularx}
    \\[5 pt] 
    *27 of 120 frames rejected
    \label{tab:t-1}
\end{table}

\vskip -10pt

\begin{table}[H]
    \caption{Roadway - Cartesian grid (no ground plane)}
    \setlength{\tabcolsep}{0.3\tabcolsep}
    \begin{tabularx}{0.48\textwidth} { 
      | >{\centering\arraybackslash}X 
      | >{\centering\arraybackslash}X 
      | >{\centering\arraybackslash}X
      | >{\centering\arraybackslash}X
      | >{\centering\arraybackslash}X
      | >{\centering\arraybackslash}X
      | >{\centering\arraybackslash}X
      | >{\centering\arraybackslash}X | }
          \hline
      & std error $x$ (cm) & std error $y$ (cm) & std error $z$ (cm) & std error $\phi$ (deg) & std error $\theta$ (deg) & std error $\psi$ (deg) \\
    \hline
    Actual** & 1.151 & 0.025 & 0.124 & 0.0057 & 0.0126 & 0.00138\\
     \hline
    Predicted & 0.437 & 0.0275 &  0.526 & 0.0236 & 0.0279 & 0.00113 \\
    \hline
    \end{tabularx}
    \\[5 pt]
    **4 of 120 frames rejected
    \label{tab:t-2}
\end{table}

\vskip -10pt

\begin{table}[H]
    \caption{Roadway - Algorithm~\ref{tab:algo}}
    \setlength{\tabcolsep}{0.3\tabcolsep}
    \begin{tabularx}{0.48\textwidth} { 
      | >{\centering\arraybackslash}X 
      | >{\centering\arraybackslash}X 
      | >{\centering\arraybackslash}X
      | >{\centering\arraybackslash}X
      | >{\centering\arraybackslash}X
      | >{\centering\arraybackslash}X
      | >{\centering\arraybackslash}X
      | >{\centering\arraybackslash}X | }
          \hline
      & std error $x$ (cm) & std error $y$ (cm) & std error $z$ (cm) & std error $\phi$ (deg) & std error $\theta$ (deg) & std error $\psi$ (deg) \\
    \hline
    Actual & 0.0958 & 0.0187 & 0.0149 & 0.00165 & 0.00159 & 0.000939\\
     \hline
    Predicted & 0.0975 & 0.0189 &  0.0131 & 0.00172 & 0.00164 & 0.00108 \\
    \hline
    \end{tabularx}
    \label{tab:t-3}
\end{table}

\vskip +5pt
\hrule
\vskip +5pt

\begin{table}[H]
    \caption{Offroad terrain - Cartesian grid}
    \setlength{\tabcolsep}{0.3\tabcolsep}
    \begin{tabularx}{0.48\textwidth} { 
      | >{\centering\arraybackslash}X 
      | >{\centering\arraybackslash}X 
      | >{\centering\arraybackslash}X
      | >{\centering\arraybackslash}X
      | >{\centering\arraybackslash}X
      | >{\centering\arraybackslash}X
      | >{\centering\arraybackslash}X
      | >{\centering\arraybackslash}X | }
     \hline
      & std error $x$ (cm) & std error $y$ (cm) & std error $z$ (cm) & std error $\phi$ (deg) & std error $\theta$ (deg) & std error $\psi$ (deg) \\
     \hline
    Actual & 0.644 & 0.253 &  0.180 & 0.00141 & 0.00735 & 0.00435 \\
    \hline
    Predicted & 0.166 & 0.117 & 0.0491 & 0.000932 & 0.00209 & 0.00236\\
    \hline
    \end{tabularx}
    \label{tab:t-4}
\end{table}

\vskip -10pt

\begin{table}[H]
    \caption{Offroad terrain - Algorithm~\ref{tab:algo}}
    \setlength{\tabcolsep}{0.3\tabcolsep}
    \begin{tabularx}{0.48\textwidth} { 
      | >{\centering\arraybackslash}X 
      | >{\centering\arraybackslash}X 
      | >{\centering\arraybackslash}X
      | >{\centering\arraybackslash}X
      | >{\centering\arraybackslash}X
      | >{\centering\arraybackslash}X
      | >{\centering\arraybackslash}X
      | >{\centering\arraybackslash}X | }
     \hline
      & std error $x$ (cm) & std error $y$ (cm) & std error $z$ (cm) & std error $\phi$ (deg) & std error $\theta$ (deg) & std error $\psi$ (deg) \\
     \hline
    Actual & 0.101  & 0.093 & 0.033 & 0.00079 & 0.00136 & 0.00162  \\
    \hline
    Predicted & 0.114 & 0.091 &  0.037 & 0.00074 & 0.00153 & 0.00186 \\
    \hline
    \end{tabularx}
    \label{tab:t-5}
\end{table}


\noindent our   spherical-grid approach is much more effective than with ground-plane removal in this case, because ground-plane removal addresses shadows that project on the ground but not those that project on the vertical walls. Not only does ground-plane removal fail to resolve vertical shadows, it also greatly increases vertical, pitch, and roll ($z,\phi,\theta$) errors, a result consistent with prior research, such as in \parencite{GP-ICP,dominantGP}, which suggested re-introducing the ground-plane specifically to compute these states.

Similar trends are observed for the $x$-errors in the off-road terrain, with the Cartesian-grid actual errors much larger than predicted errors (by a factor of four) and with the spherical-grid actual errors close to predicted errors (within 13\%). 
For the offroad scene, the actual errors in the other directions ($y,z,\phi,\theta,\psi$) were also larger for the Cartesian grid, by a factor of 2-3 times, as compared to prediction; our new algorithm again predicted the actual errors well in these directions (within 15\%).

 For both scenes, perhaps surprisingly, the spherical grid resulted in higher accuracy than the Cartesian grid. We hypothesize this results from better conditioning due to more consistent clustering of points in each voxel, noting that the number of Lidar points varies little across the voxels on the spherical grid but varies substantially across the voxels in the Cartesian grid, as seen in Fig.~\ref{fig:scene2_viz}.

\section{Discussion}

\begin{figure}
\centering
\includegraphics[width=1.5in]{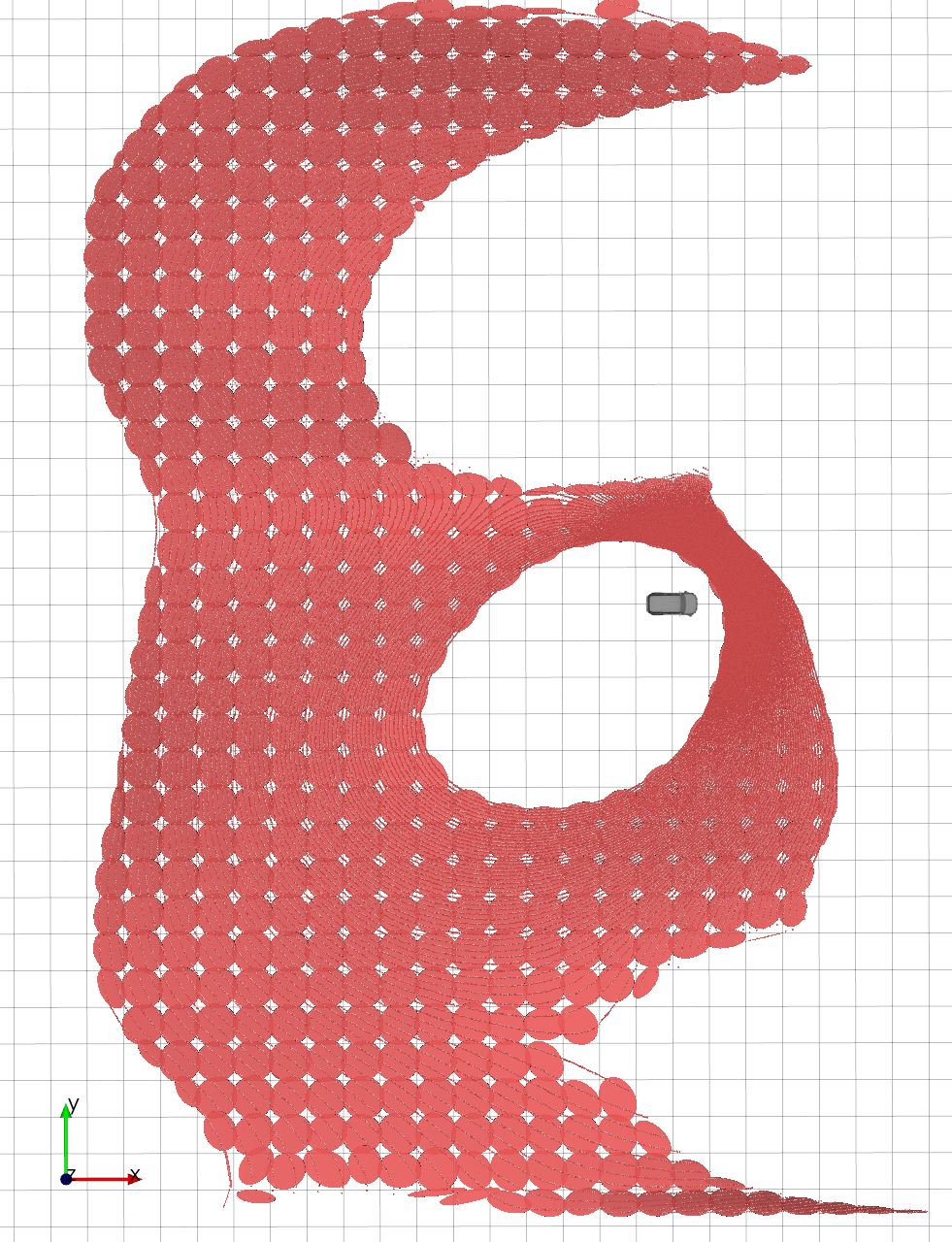}
\includegraphics[width=1.45in]{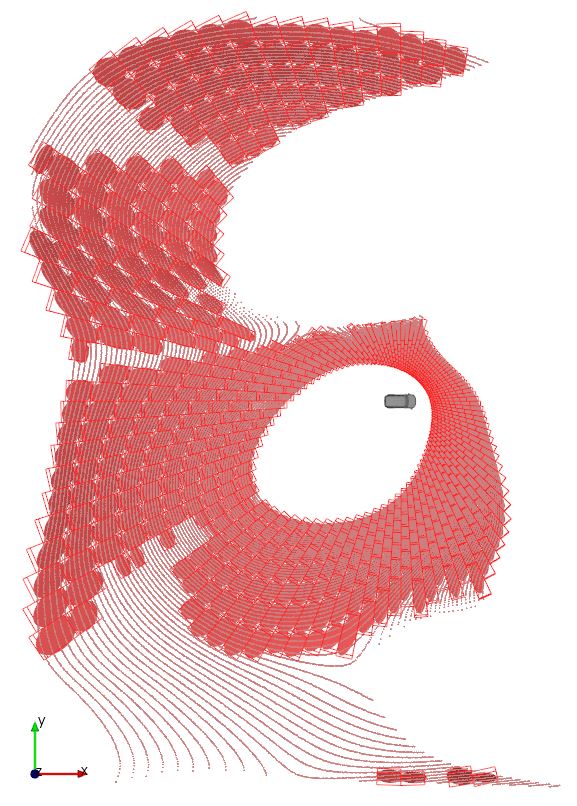}
\caption{Overhead view of Cartesian (left) and spherical (right) grids for offroad scene, with distribution ellipsoids (red) shown for each voxel used in the scan-matching computation}
\label{fig:scene2_viz}
\end{figure}

As demonstrated by Tables~\ref{tab:t-1}-\ref{tab:t-5}, our proposed shadow-mitigation strategy improves scan-matching accuracy by using a spherical voxel grid instead of a Cartesian one. The spherical voxels also greatly enhance the quality of accuracy predictions made by ICET in the presence of significant shadowing. As discussed above, other benefits of the spherical grid include reduced computational effort (due to the use of a 2D grid in place of a 3D grid) and less sensitivity to grid placement. Grid sensitivity can be visualized at the edges of the left image in Fig.~\ref{fig:scene2_viz}, where ellipsoids are poorly defined because relative few scan points appear in voxels far from the Lidar.

Despite the many advantages of the spherical grid introduced as part of Algorithm~\ref{tab:algo}, there are likely to be several limitations, which we will explore in future work.
We suspect one disadvantage of our spherical-grid approach is that existing search-acceleration strategies \parencite{EINHORN201528} will work poorly for the case of large initial uncertainty, such as in the lost robot problem where initial conditions are not known. Another limitation is that our method only addresses errors caused by shadowing. The proposed algorithm does not address other error sources (such as moving objects and perspective shifts), and these anomalies would be expected to cause differences between true and predicted \mbox{$\sigma$-values} for error in many scenes.

\section{Conclusion}

This paper introduced a new method for mitigating shadowing errors in Lidar scan matching. This shadow-matching algorithm runs as a preprocessing step before scan matching by identifying the nearest meaningful cluster of points in a set of azimuth and elevation bins, in effect creating a 2D voxel grid in spherical coordinates. This grid aligns with Lidar rays, and with shadow edges; thus, shadow edges can be eliminated by preserving only the nearest point cluster in each wedge-shaped azimuth/elevation bin. As compared to the commonly used technique of ground-plane removal, our proposed method offers advantages in that it works even on oblique or uneven terrain and in that it retains useful information needed to compute vertical, pitch, and roll states accurately. For voxel-based scan matching methods like NDT or ICET, the resulting grid can be reused by the scan matching algorithm. The 2D spherical grid offers several advantages over a 3D Cartesian grid for scan matching including reduced computational costs and better numerical conditioning due to more uniform balancing of the number of points in each voxel. These conclusions were verified through Monte Carlo simulations of urban and offroad terrains.


 
\section{Acknowledgements}
The authors wish to acknowledge and thank the U.S. Department of Transportation Joint Program Office (ITS JPO) and the Office of the Assistant Secretary for Research and Technology (OST-R) for sponsorship of this work. We also gratefully acknowledge NSF grant CNS-1836942, which supported specific aspects of this research. Opinions discussed here are those of the authors and do not necessarily represent those of the DOT, NSF, or other affiliated agencies. 

\printbibliography

\end{document}